\documentclass[sigconf]{acmart}
\usepackage{balance}
\AtBeginDocument{%
  }

\copyrightyear{2026}
\acmYear{2026}
\setcopyright{cc}
\setcctype{by}
\acmConference[MM '26]{Proceedings of the 34th ACM International Conference on Multimedia}{November 10--14, 2026}{Rio de Janeiro, Brazil}
\acmBooktitle{Proceedings of the 34th ACM International Conference on Multimedia (MM '26), November 10--14, 2026, Rio de Janeiro, Brazil}
\acmDOI{10.1145/3767308.3836446}
\acmISBN{979-8-4007-2213-4/2026/11}
\begin{document}



\title{How Do VLMs Fail? Vision-Operation Misalignment in Compositional VQA
}
% How Do VLMs Fail? Vision–Operation Misalignment in Compositional VQA

\author{Navya Gupta}
\email{2570043@sit.singaporetech.edu.sg}
\affiliation{%
  \institution{Singapore Institute of Technology}
  \city{Singapore}
  \country{Singapore}
}

\author{Bingjie Xu}
\email{bingjie.xu@singaporetech.edu.sg}
\affiliation{%
  \institution{Singapore Institute of Technology}
  \city{Singapore}
  \country{Singapore}
}

\author{Avinash Anand}
\email{avinash.anand@singaporetech.edu.sg}
\affiliation{%
  \institution{Singapore Institute of Technology}
  \city{Singapore}
  \country{Singapore}
}

\author{Timothy Liu}
\email{timothyl@nvidia.com}
\affiliation{%
  \institution{NVIDIA}
  \city{Singapore}
  \country{Singapore}
}

\author{Zhengchen Zhang}
\authornote{Corresponding author.}
\email{zhengchen.zhang@singaporetech.edu.sg}
\affiliation{%
  \institution{Singapore Institute of Technology}
  \city{Singapore}
  \country{Singapore}
}

\renewcommand{\shortauthors}{Navya Gupta, Bingjie Xu, Avinash Anand, Timothy Liu, \& Zhengchen Zhang}
%% No italics, no superscripts, not anonymous
%% Use footnote or author note to identify equal contribution, shared contribution, and/or contact author info

\begin{abstract}
Compositional visual question answering requires Vision-Language Models (VLMs) to execute multiple reasoning operations like object selection, spatial relation resolution, and attribute verification. Despite strong aggregate performance, the mechanistic basis of VLM failures on this task remains underexplored.
To address this gap, we analyze vision–operation misalignment in VLMs by examining how failures relate to specific reasoning operations and the internal computational pathways through which they arise and propagate.
We introduce an \textit{Operation-centric mechanistic framework} that decomposes VLM failures by both the reasoning operation where they originate and the internal computational pathway through which they propagate. 
Our analysis reveals four dominant failure modes: grounding failure, reasoning failure, attribute extraction failure, and language-prior dominance, each characterized by a distinct relationship between visual grounding strength and answer correctness. Through three complementary causal interventions applied across all transformer layers, we find that object-selection failures are associated primarily with feedforward computation, multi-step relational failures with late-layer direct attention, and attribute-extraction failures with answer-position feedforward computation. Validation on VSR further shows that single-step spatial failures are concentrated at object-position encoding, distinguishing them from multi-step relational composition. These findings reveal distinct computational bottlenecks across operation types and provide a principled basis for targeted diagnosis of VLM failures in multimedia reasoning.
\end{abstract}





\maketitle

\section{Introduction}

Compositional visual question answering requires chaining multiple reasoning steps over grounded visual evidence like localizing objects, evaluating attributes, check for existence, and computing spatial relations. On benchmarks designed to test this capability, such as GQA~\cite{hudson2019gqa}, modern Vision-language models(VLMs)~\cite{bai2025qwen3} report strong aggregate accuracy. Yet aggregate performance conceals systematic failure patterns: models succeed through linguistic shortcuts on easy operations while failing on those that genuinely require visual grounding and multi-step inference. Understanding these failures requires moving beyond what the model outputs to examining how visual and linguistic computation interact at each reasoning step.

\begin{figure}[t]
\centering
\includegraphics[width=\columnwidth]{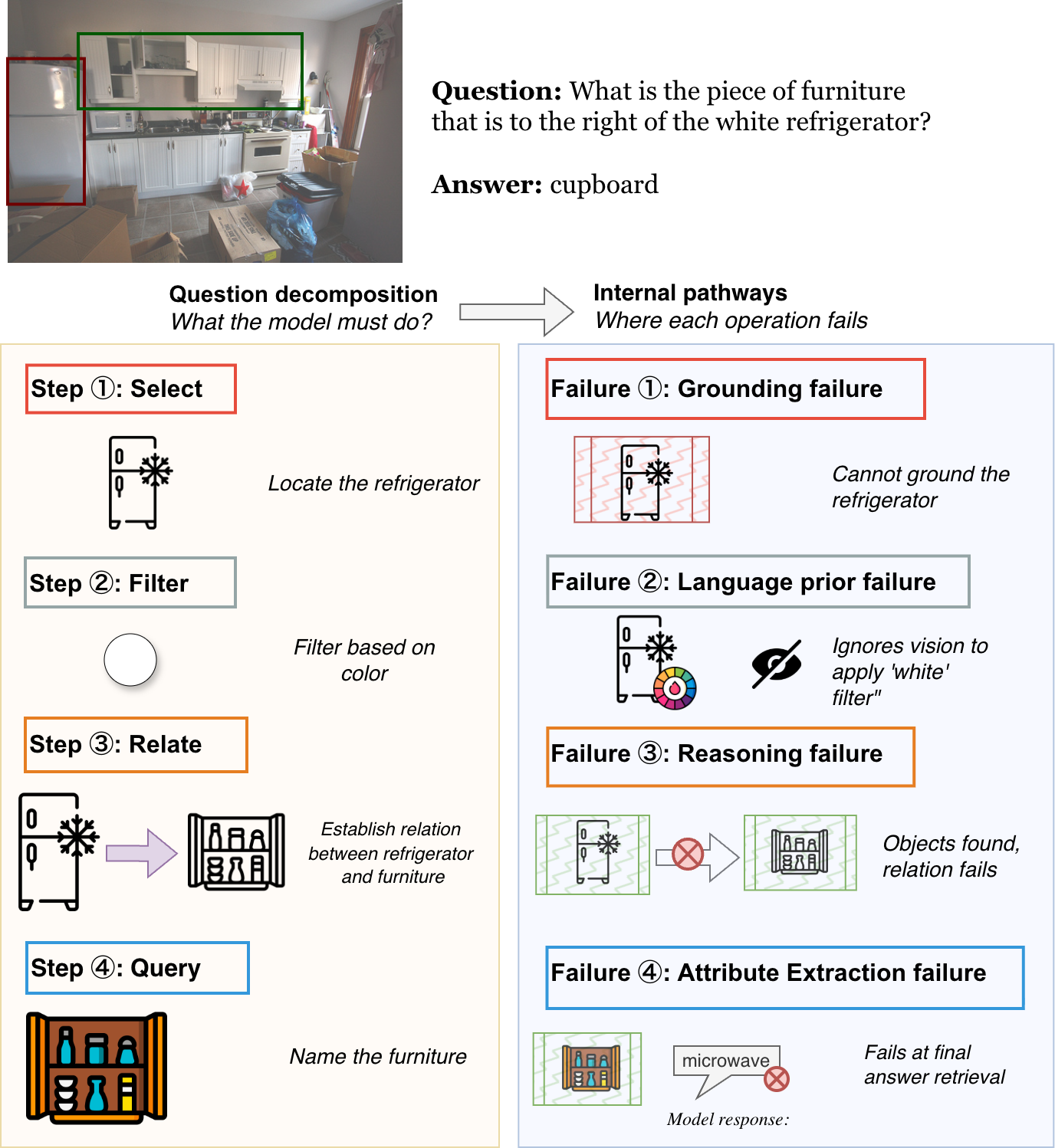}
\caption{\textbf{Four failure modes in compositional VQA, each 
breaking at a different internal stage.} A chained question can fail at perception, reasoning, or answer retrieval. Fourth failure mode is language prior dominance which relies predominantly on language priors.}
\Description{Four stages at which a compositional VQA question can fail: object grounding, relational reasoning, attribute extraction, and language-prior dominance.}
\label{fig:motivation}
\end{figure}

Consider the question ``What is the piece of furniture that is to the right of the white refrigerator?'' Answering correctly requires four operations: selecting the refrigerator, filter based on the color, resolving the spatial relation to identify the adjacent object, and answer the query regarding name of the furniture(See Figure~\ref{fig:motivation}). Each operation demands a different kind of vision-language integration, and a model can fail at any of them. Critically, the failure mechanism differs by operation type and failing to locate an object is a fundamentally different computational problem from locating it correctly but misjudging a spatial relation. No existing analysis framework distinguishes these failure types at the mechanistic level.

In transformer-based VLMs, visual information reaches the answer through two pathways: cross-token attention routing, where the answer token reads from object positions via the attention mechanism, and within-token feedforward transformation, where the MLP network processes and transforms representations at each position independently. Which pathway carries the visual signal for which operation, and where that signal is lost when the model fails, are open questions that behavioral evaluation cannot answer.

Mechanistic interpretability offers the tools to answer them. Intervention techniques such as causal ablation ~\cite{meng2022locating, neo2024towards} and attention knockout ~\cite{geva-etal-2023-dissecting} have been applied to language models to identify which attention heads and layers carry specific computations. Linear probing ~\cite{belinkov-2022-probing,ferrando-voita-2024-information} provides complementary representational evidence. However, existing VLM interpretability work~\cite{palit2023towards, ben2024lvlm} has largely focused on broad failure types, typically object hallucination~\cite{li2025causal} and cross-modal information flow~\cite{Zhang_2025_CVPR, basu2024understanding}. Existing analyses identify visual, grounding, and reasoning weaknesses, but do not jointly diagnose how failures associated with different compositional operations relate to distinct internal computational loci. 
We address this question through an \textit{operation-centric mechanistic framework} that decomposes VLM failures by the typed reasoning operation where they occur and the internal computational pathway associated with their occurrence.

Leveraging GQA's functional program annotations and scene graph bounding boxes, we identify the specific object visual tokens causally relevant to answer the question, not ablating all vision tokens uniformly.
Each question is decomposed into operations (\texttt{select}, \texttt{relate}, \texttt{verify}, \texttt{query}, \texttt{exist}, \texttt{choose}, \texttt{filter}).

We identify four mechanistically distinct failure modes that differ both in the stage of the forward pass where they arise and in the internal pathway through which they operate: \textbf{Grounding failure}, \textbf{Reasoning failure}, \textbf{Attribute extraction failure}, and \textbf{Language prior dominance}

The key mechanistic insight is that different failure modes have different dominant computational loci. Object-selection and attribute-extraction failures are associated primarily with feedforward computation, whereas multi-step relational failures on GQA are associated with late-layer direct attention to grounded objects. VSR further shows that single-step spatial failures are concentrated at object-position MLP computation. These distinctions provide a basis for operation-aware diagnosis of VLM reliability in VQA and multimedia reasoning.

Our contributions are:
\begin{enumerate}
    \item \textbf{An operation-aware failure taxonomy for 
    compositional VQA.} We decompose VLM failures into four 
    mechanistically distinct modes: grounding failure, 
    reasoning failure, attribute extraction failure, and 
    language prior dominance failure. Each defined by a 
    unique relationship between visual grounding strength 
    and answer correctness across seven GQA operation types. 
    Cross-architecture analysis on LLaVA-1.5-7B~\cite{llava} shows that attribute-extraction effects transfer most clearly, while grounding and relational profiles are more sensitive to the vision encoder. Image-corruption controls independently support language-prior dominance, although its precise intervention profile varies across models.

    \item \textbf{Operation-specific causal targeting via 
    scene-graph grounding.} Rather than intervening on all 
    visual tokens uniformly, we identify the specific vision 
    tokens causally relevant to each reasoning operation 
    using GQA's functional programs and scene graphs. This 
    operation-aware targeting reveals that grounding strength 
    predicts correctness differently across operations --- 
    positively for object selection, inversely for spatial 
    reasoning, and not at all for language-prior operations.

    \item \textbf{Distinct computational loci across failure modes.}
By separately targeting direct attention and feedforward computation, we find that object-selection and attribute-extraction failures are associated primarily with feedforward processing, while multi-step relational failures on GQA are associated with late-layer direct attention. VSR further distinguishes single-step spatial failures, which are concentrated at object-position encoding.

\end{enumerate}

\section{Related Work}

\paragraph{Mechanistic interpretability in language and vision-language models.}
In language models, this line of work has established a set of causal and representational 
tools for analysing internal computation. Attention heads have been 
characterised as structured circuits with distinct query-key (QK) 
and output-value (OV) roles ~\cite{elhage2021mathematical}, while 
targeted interventions enable localisation of task-relevant information within 
specific layers and pathways ~\cite{geva-etal-2023-dissecting, meng2022locating, zhang2024towards}. 
Complementary methods including residual stream decomposition and 
linear probing provide additional insight into how representations 
are formed, transformed, and propagated across depth 
~\cite{ferrando-voita-2024-information, alain2016understanding, belinkov-2022-probing}. 
Recent work extends these techniques to VLMs, 
adapting causal tracing and perturbation-based analysis to study 
cross-modal computation ~\cite{lin2025survey, palit2023towards, golovanevsky2024vlms, liu2025mechanisticinterpretabilitymeets}. 
These studies show that visual information is progressively integrated 
into linguistic representations, with object-level signals emerging and 
refining across layers ~\cite{neo2024towards, yu2024understanding}, and 
cross-modal interaction occurring through structured routing mechanisms 
such as specialised attention heads and staged integration pipelines 
~\cite{Zhang_2025_CVPR, kim2025interpreting}. Further work demonstrates 
that object representations can be causally localised and linked to model 
predictions, particularly in the context of hallucination and grounding 
errors ~\cite{li2025causal}. 

\paragraph{Attention and MLP pathway analysis.}
In language models, attention mechanisms primarily route 
information across tokens while MLP layers act as 
key-value memories storing factual 
associations~\cite{elhage2021mathematical, 
geva-etal-2021-transformer, meng2022locating}. 
Recent work extends this to multimodal models, revealing 
that visual constraints are encoded in early MLP and 
self-attention blocks while mid-layer attention routes 
this information to the output~\cite{basu2024understanding}, and certain attention heads specialise in reasoning-related functions~\cite{jiang2025investigating}. Other studies trace how visual capabilities emerge across 
layers during fine-tuning~\cite{naghashyar2026towards}.However, the relationship between these pathways and specific failure modes in compositional VQA remains unexplored.

\paragraph{Failure analysis in VLMs.}
Understanding why VLMs fail on visual question answering has been approached from behavioral and representational perspectives. 
Object hallucination has been extensively studied as a language-prior 
failure, where answers are driven by statistical co-occurrence rather 
than visual evidence ~\cite{rohrbach-etal-2018-object, li2023evaluating, luo2024probing}. 
Attribute binding failures, where models ground the correct object 
but retrieve the wrong property, are documented through compositional 
benchmarks such as Winoground ~\cite{thrush2022winoground} and VALSE ~\cite{parcalabescu2022valse}. Spatial and relational failures are characterised in VSR ~\cite{liu2023visual, tong2024eyes}, showing that models 
systematically fail at positional resolution even when 
objects are correctly localised. At a higher level, VLMs have been shown to underperform 
their own vision encoders on basic image classification ~\cite{zhang2024visually}, and critic-based 
frameworks that decouple reasoning from verification can 
partially mitigate such errors through iterative 
refinement~\cite{zhang2025critic}. 
%However, these analyses treat failure as a behavioral phenomenon without connecting it to specific internal computational pathways. 

\begin{figure*}[ht]
  \includegraphics[width=\textwidth]{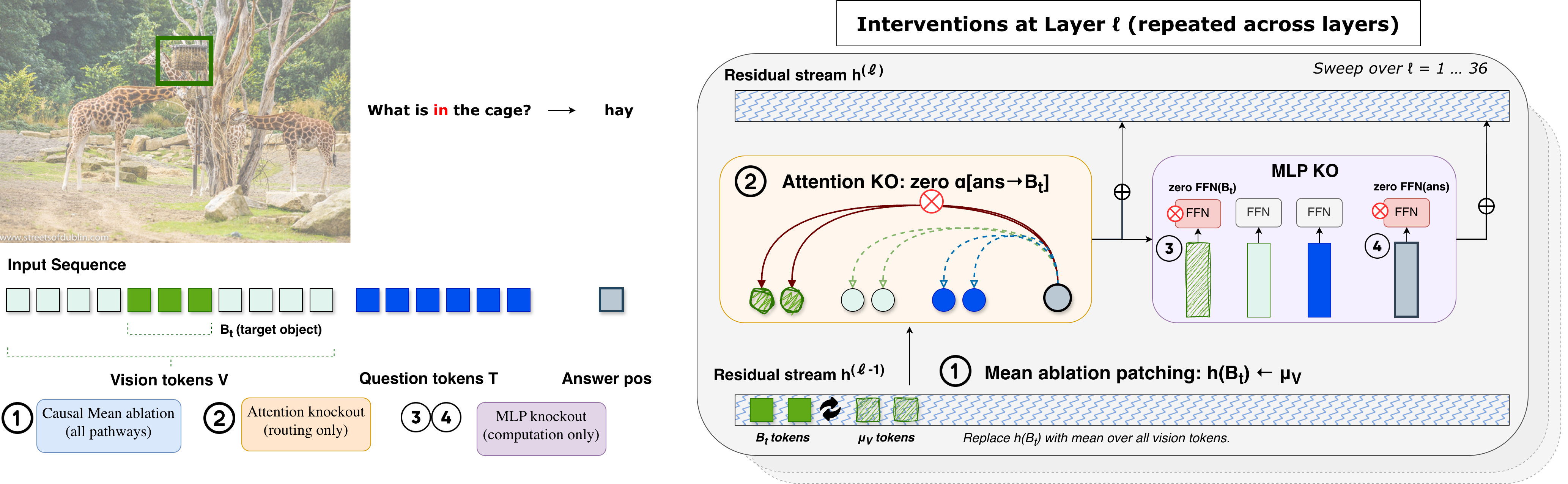}
  \caption{
\textbf{Layer-wise causal interventions for object-specific reasoning.}
At layer $\ell$, we intervene on target object tokens $B_t$ using three methods: Causal Mean ablation ($h(B_t) \leftarrow \mu_V$), attention knockout ($\alpha_{\text{ans}\rightarrow B_t}=0$), and MLP knockout (zeroing FFN outputs at object and answer positions).
}
\Description{Layer-wise interventions on target object tokens:
mean replacement, blocking answer-to-object attention, and
zeroing MLP outputs at object or answer positions.}
  \label{fig:method}
\end{figure*}

\section{Methodology}

\subsection{Preliminaries}

We study Qwen2.5-VL-3B-Instruct ~\cite{bai2025qwen3}, a 
vision-language model consisting of a native-resolution 
vision encoder and a 36-layer transformer language 
backbone. The input sequence 

$\mathbf{x} = [v_1, \dots, v_{N_V}, t_1, \dots, t_{N_T}]$ 

concatenates $N_V$ vision tokens from the image encoder 
with $N_T$ text tokens from the tokenized question, 
followed by a single answer position from which the 
model generates its response.

Each transformer layer $\ell$ updates the hidden state 
at position $i$ in two sequential steps:
\begin{align}
\hat{H}_i^{(\ell)} &= H_i^{(\ell-1)} + 
\mathrm{Attn}\!\left(\mathrm{LN}\!\left(H_i^{(\ell-1)}\right)\right) \\
H_i^{(\ell)} &= \hat{H}_i^{(\ell)} + 
\mathrm{MLP}\!\left(\mathrm{LN}\!\left(\hat{H}_i^{(\ell)}\right)\right)
\end{align}
where $\mathrm{LN}$ denotes layer normalisation. The 
attention output decomposes over heads $h$ as 
$a_i^{(\ell)} = \sum_h \sum_j 
\alpha_{ij}^{(\ell,h)} W_O^{(\ell,h)} W_V^{(\ell,h)} 
H_j^{(\ell-1)}$, where $\alpha_{ij}^{(\ell,h)}$ are 
post-softmax attention weights, and $W_V^{(\ell,h)}$, 
$W_O^{(\ell,h)}$ are the value and output projection 
matrices for head $h$.

We denote the set of vision token positions as 
$\mathcal{V}$, the subset corresponding to a target 
object's bounding box as $\mathcal{B}_t$, text positions 
as $\mathcal{T}$, and the answer position as 
$\mathrm{ans}$. All interventions measure degradation in 
the log-probability of the correct answer: positive 
degradation indicates the intervened component was 
causally supporting the correct prediction.

\begin{table}[t]
\centering
\caption{\textbf{Qualitative examples by operation and failure mode.}
Correct (\checkmark) and incorrect ($\times$) predictions drawn from GQA}
\label{tab:qualitative}
\small
\setlength{\tabcolsep}{4pt}
\renewcommand{\arraystretch}{1.3}
\begin{tabular}{l p{4cm} p{1.0cm} p{1.2cm}}
\toprule
\textbf{Op.} & \textbf{Question} & \textbf{GT} & \textbf{Pred.} \\
\midrule

\multicolumn{4}{l}{\textit{Grounding failure}} \\
select$\times$ & Does the helmet have a different color than the seat? & no & yes \\
select\checkmark & Is the bench made of the same material as the scaffolding? & yes & yes  \\
\midrule

\multicolumn{4}{l}{\textit{Reasoning failure}} \\
relate$\times$ & Where is the girl? & beach & water \\
relate\checkmark & Which kind of furniture is the man \textbf{on}? & bed & bed  \\
\midrule

\multicolumn{4}{l}{\textit{Attribute extraction failure}} \\
verify$\times$ & Is the boy holding the stick? & yes & no \\
query$\times$  & What color is the cart? & golden & silver  \\
exist\checkmark  & Are there both plates and spoons in the picture? & yes & yes  \\
\midrule

\multicolumn{4}{l}{\textit{Language prior}} \\
filter\checkmark & Is the bottle that is made of glass open or closed? & open & open\\
choose\checkmark & Is the chain black or yellow? & black & black \\
\bottomrule
\end{tabular}
\end{table}

\subsection{Operation-Aware Decomposition of Queries and Failure Taxonomy}

GQA annotates each question with a functional program $\mathcal{P}(q) = (o_1, \dots, o_T)$ that decomposes it into typed reasoning steps, and a scene graph that grounds each step's arguments to bounding box regions in the image. 
The operation vocabulary includes \texttt{select} (locate 
an object), \texttt{relate} (resolve a spatial relation), 
\texttt{verify} (check a property or relation), 
\texttt{query} (retrieve an attribute), \texttt{exist} 
(test presence), \texttt{filter} (subset by attribute), 
and \texttt{choose} (select between alternatives).

For example, the question \textit{``Is the object to the 
left of the red chair made of wood?''} decomposes into 
three operations: \texttt{select}(red chair) $\to$ 
\texttt{relate}(to the left of) $\to$ 
\texttt{verify}(made of wood). Each operation is grounded 
to a specific image region through the scene graph, and 
the model must integrate the correct visual evidence at 
each step to answer correctly. Our analysis asks whether 
failure at each step has a distinct mechanistic signature.

\paragraph{Operation-aware grounding}
Each operation $o_t$ is associated with bounding box 
regions $\mathcal{B}_t$ resolved from the scene graph. 
We define grounding sets based on causal relevance to the 
operation rather than naively patching all mentioned 
objects:
\begin{itemize}
\item \texttt{select}, \texttt{query}, \texttt{filter}: 
  the selected or filtered object
\item \texttt{exist}: each branch object independently
\item \texttt{verify}, \texttt{choose}: primary object and 
  relational argument jointly, since the relation requires 
  both reference points
\item \texttt{relate}: the terminal relate-step object 
  only which determines the answer
\end{itemize}

This reflects causal structure for \texttt{verify}, patching either reference object alone yields incomplete signal. For \texttt{relate}, intermediate objects are not answer-determining and introduce noise. 
These grounding sets $\mathcal{B}_t$ are shared across all 
subsequent experiments.

\paragraph{Four failure modes.}
Comparing grounding strength between correct and incorrect predictions reveals four dominant failure modes. Table \ref{tab:qualitative} shows qualitative examples to shows failure modes of each operation. \textbf{Grounding failure} occurs when correct predictions depend more strongly on the target region than incorrect predictions, indicating unsuccessful object localization or encoding. \textbf{Reasoning failure} shows the reverse relationship: incorrect predictions depend more strongly on the grounded object, indicating over-grounding during relational processing. \textbf{Attribute extraction failure} exhibits strong grounding regardless of correctness, but fails when converting the grounded representation into an answer. \textbf{Language-prior dominance} exhibits weak visual dependence for both correct and incorrect predictions. These modes summarize dominant operation-level patterns.

\subsection{Measuring Visual Criticality via Causal Mean 
Ablation}

To determine whether visual grounding is causally necessary for each operation, we ablate the operation's grounding tokens $\mathcal{B}_t$ at each layer $\ell$ by replacing their hidden states with a neutral baseline, and measure the resulting degradation in the log-probability of the ground-truth answer $a$ given image $I$ and question $q$:
\begin{equation}
\Delta_t^{(\ell)} = \log p(a \mid I, q) - \log 
p\!\left(a \mid I, q;\, 
\mathrm{ablate}(\mathcal{B}_t, \ell)\right)
\end{equation}
The baseline is the mean activation over all vision token positions at layer $\ell$ from the clean forward pass:
\begin{equation}
\mu^{(\ell)} = \frac{1}{|\mathcal{V}|} \sum_{j \in 
\mathcal{V}} H_j^{(\ell)}
\end{equation}
We use the per-sample mean over vision tokens as the replacement baseline because it is in-distribution and denotes a valid point in the activation space the model has 
processed. It also preserves the aggregate visual context 
while removing only object-specific information, ensuring 
that degradation reflects the loss of the target object 
rather than the removal of all visual signal.

Crucially, only the tokens in $\mathcal{B}_t$ are replaced, the remaining vision tokens and all text tokens are untouched. This removes object-specific information required by the model to answer the question while keeping activations in-distribution, and ensures that the measured degradation reflects the causal role of the specific grounded object, not of global visual input. Our intervention tests necessity, whether the model requires the object's representation at layer $\ell$ to produce the correct answer.

The degradation curve $\{\Delta_t^{(\ell)}\}_{\ell=1}^{L}$ characterises how causal reliance on the object evolves across network 
depth. We represent it with the \emph{grounding strength} $\mathrm{GS}(o_t) = \max_\ell \Delta_t^{(\ell)}$, capturing how much the operation depends on visual grounding.

As a baseline-sensitivity control, we repeat the analysis using zero-vector replacement, $H_j^{(\ell)}\leftarrow\mathbf{0}$ for $j\in\mathcal{B}_t$.

\subsection{Isolating Computational Pathways via Targeted 
Knockout}
\label{sec:knockout}

Causal mean ablation removes the object's hidden state entirely, affecting both the attention pathway (other tokens reading from the object position) and the feedforward pathway (MLP transformations of the object and answer representations). To test the pathway hypotheses, we apply two complementary interventions that isolate each pathway independently(See Figure~\ref{fig:method}).

\paragraph{Attention knockout.}
To isolate the attention pathway, we prevent the answer token from attending to the object's visual tokens while preserving the remaining attention structure. At layer $\ell$, attention weights from the answer position to grounding tokens $\mathcal{B}_t$ are set to zero and the remaining weights renormalized:
\begin{equation}
\tilde{\alpha}^{(\ell,h)}_{\mathrm{ans},j} =
\begin{cases}
0 & j \in \mathcal{B}_t \\[4pt]
\dfrac{\alpha^{(\ell,h)}_{\mathrm{ans},j}}
{1 - \sum_{k \in \mathcal{B}_t} 
\alpha^{(\ell,h)}_{\mathrm{ans},k}} 
& j \notin \mathcal{B}_t
\end{cases}
\end{equation}
Renormalization ensures the intervention changes only \emph{where} each head reads, not how much --- the 
value-weighted output retains its original scale. The 
knockout degradation $\mathrm{KO}_{\mathrm{attn}}^{(\ell)} = \log p(a \mid I, q) - \log p(a \mid I, q;\, \tilde{\alpha}^{(\ell)})$ measures causal dependence on attention-mediated object access at each layer.

This intervention blocks only the \emph{direct} attention path from the answer token to object positions at layer $\ell$. Indirect paths are not blocked. This is an intentional design choice as blocking all indirect paths would require intervening across all layers simultaneously, confounding the layer-wise localisation that is the primary goal of the experiment. The layer-wise degradation curve therefore measures the marginal causal contribution of direct attention routing at each depth, not total information flow from object tokens.
\paragraph{MLP knockout.}
To isolate the feedforward pathway, we zero out the MLP output at specific token positions. We consider two variants targeting different computational roles:
\begin{equation}
\tilde{f}^{(\ell)}_i = 
\begin{cases}
\mathbf{0} & i \in \mathcal{S} \\[4pt]
f^{(\ell)}_i & i \notin \mathcal{S}
\end{cases}
\end{equation}
where $\mathcal{S}$ is the set of target positions and $f^{(\ell)}_i$ is the MLP output at position $i$, layer $\ell$. In \emph{object-position knockout} ($\mathcal{S} = \mathcal{B}_t$), the MLP cannot transform the object's representation, disrupting within-token visual processing while leaving attention routing intact. In \emph{answer-position knockout} ($\mathcal{S} = \{\mathrm{ans}\}$), the MLP cannot transform the answer token's representation, disrupting the computation that converts attended information into the final prediction.

\paragraph{Pathway comparison.}
The three degradation curves: Causal Mean ablation 
$\Delta^{(\ell)}$, Attention knockout $\mathrm{KO}_{\mathrm{attn}}^{(\ell)}$, and MLP knockout $\mathrm{KO}_{\mathrm{mlp}}^{(\ell)}$ are computed on the same samples with the same correctness labels, enabling direct comparison. Stratifying each degradation metric by model correctness and computing the grounding specificity shift (Cohen's $d$) identifies which targeted computational component most strongly distinguishes correct from incorrect predictions for each operation. The joint signature across all three metrics assigns each operation to its failure mode 
(Table~\ref{tab:main-results}).

\section{Experimental Setup}

\paragraph{Data and sampling.}
We use GQA ~\cite{hudson2019gqa}, which provides 
compositional questions paired with functional programs 
and scene graphs over real images. Scene graphs supply 
bounding boxes for all grounded objects, enabling 
operation-aware identification of the vision tokens 
$\mathcal{B}_t$ relevant to each reasoning step. We 
sample 500 questions per operation type (\texttt{select}, 
\texttt{relate}, \texttt{verify}, \texttt{exist}, 
\texttt{query}, \texttt{choose}, \texttt{filter}) from 
the balanced training split for 3{,}500 total samples. 
Generalisation experiments on VSR ~\cite{liu2023visual} for \texttt{relate} operation are reported in 
\S\ref{sec:results}.

\paragraph{Model and inference.}
All experiments use Qwen2.5-VL-3B-Instruct 
~\cite{bai2025qwen3} in fp16 precision. Model correctness 
is determined by greedy generation with 
matching against the ground truth. Answer log-probabilities are computed by teacher-forcing the ground truth tokens. 
All interventions are applied one layer at a time across all 36 
layers, producing a degradation curve per sample. Each 
experiment requires 37 forward passes per sample (1 clean 
+ 36 intervened).

\paragraph{Metrics and statistics.}
We report three quantities per operation, each split by 
model correctness:
\begin{itemize}
\item \textbf{Grounding strength} (GS): peak degradation 
  under mean ablation across layers, measuring total 
  causal dependence on the object's visual representation.
\item \textbf{Attention knockout degradation} 
  ($\mathrm{KO}_\mathrm{attn}$): peak degradation when 
  attention from the answer position to $\mathcal{B}_t$ 
  is blocked, isolating the attention pathway.
\item \textbf{MLP knockout degradation} 
  ($\mathrm{KO}_\mathrm{mlp}$): peak degradation when the 
  MLP output at the answer position is zeroed, isolating 
  the feedforward pathway.
\item \textbf{Grounding specificity shift} (GSS): Cohen's 
  $d$ between correct and incorrect samples on each 
  degradation metric, serving as the primary statistic 
  for failure mode classification:
  Positive GSS indicates correct predictions show higher 
  degradation (grounding aids success); negative GSS 
  indicates incorrect predictions show higher degradation 
  (over-grounding).
\end{itemize}

Each failure mode is identified by its joint signature across all three 
metrics (Table~\ref{tab:main-results}).

\paragraph{Representational validation.}
To support that causal findings reflect visual computation rather than surface text features, we construct contrastive question pairs $(q_\text{clean}, q_\text{corrupt})$ from GQA that share the same image but differ in a semantically relevant word (e.g.\ replacing an object name or attribute). We extract hidden states at each layer for both members of the pair and train a logistic regression probe to classify each state as clean or corrupted. 
Pairs are partitioned by \emph{behavioural vision sensitivity}: a pair is vision-sensitive (VS) if the model's generated answer changes between the clean and corrupted input, and vision-insensitive (NV) otherwise (Eq.~\ref{eq:vs}).
\begin{equation}\label{eq:vs}
\text{VS}(q) = \mathbf{1}\!\left[f(I, q_\text{clean}) 
\neq f(I, q_\text{corrupt})\right]
\end{equation}
The probe accuracy gap $\delta_\text{VS} = \text{acc}_\text{VS} - \text{acc}_\text{NV}$ serves as a validity criterion. We require $\delta_\text{VS} > 0.05$ to treat an operation's causal results as reflecting vision-grounded representations rather than text artifacts.

\paragraph{Spatial relation generalization.}
The GQA analysis decomposes failures across seven operation 
types, but each question involves multiple chained 
operations, making it difficult to isolate the spatial 
reasoning component from upstream grounding steps. To 
disentangle this, we replicate all three interventions on 
VSR~\cite{liu2023visual}, a binary spatial reasoning 
benchmark where each question requires a single spatial 
judgment between two objects (e.g.\ \emph{``The cat is 
behind the laptop''}; answer yes/no). We use the test 
split (2{,}195 samples, 66 relation types across 7 
meta-categories), applying 
the same protocol as GQA and intervening on the subject, 
object, and both jointly.

\section{Results}\label{sec:results}
Table~\ref{tab:main-results} presents the complete mechanistic profile across three interventions. Each intervention is applied independently at every layer, and we report the peak degradation in answer log-probability, split by model correctness, with Cohen's $d$ quantifying the correctness--degradation association. As a robustness check on peak-based aggregation, we also compute AUC across all 36 layers; the resulting operation ranking closely matches the peak-based ranking (Spearman $\rho=0.96$, $p=0.0005$). 
As a baseline-sensitivity control, zero-vector replacement preserves significant correctness-dependent effects for \texttt{relate} ($|d|=0.35$, $p<0.001$) and \texttt{exist} ($|d|=0.39$, $p=0.001$), while \texttt{filter} remains non-significant ($|d|=0.12$, $p=0.44$). \texttt{choose} shows a nominal effect ($|d|=0.43$, $p=0.04$), indicating that the control does not reproduce every signed operation profile identically. The image-corruption control provides independent support for the language-prior interpretation.

The joint pattern across all three interventions assigns each operation to one of four failure modes.

\begin{figure}[t]
\centering
\includegraphics[width=\columnwidth]{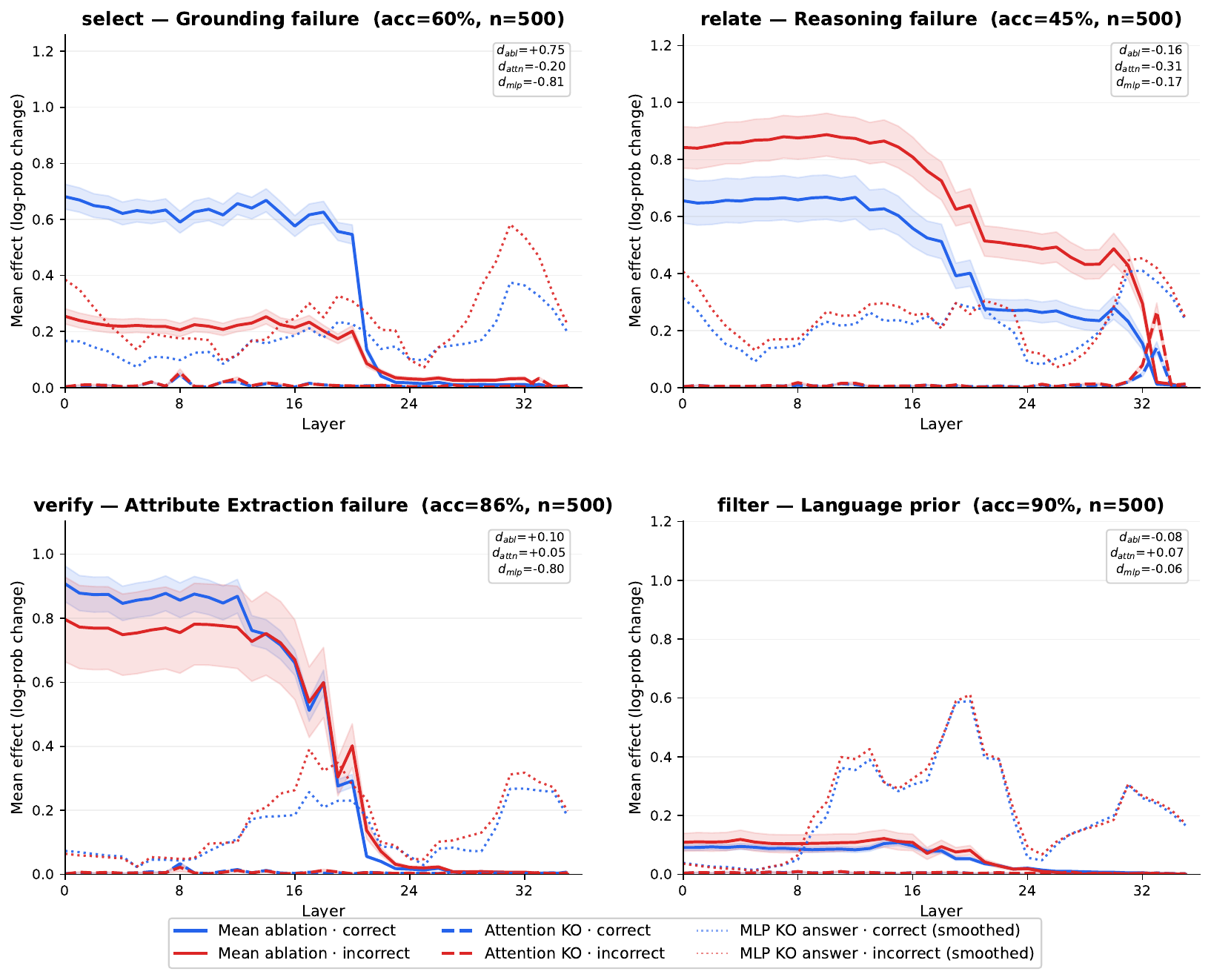}
\caption{\textbf{Four mechanistically distinct failure modes.}
Layer-wise degradation under mean ablation (solid), attention 
knockout (dashed), and MLP knockout (dotted), split by correctness 
(blue = correct, red = incorrect). 
}
\Description{Four panels comparing layer-wise degradation for
correct and incorrect predictions under mean ablation, attention
knockout, and MLP knockout for the four failure modes.}
\label{fig:pathway-dissociation}
\end{figure}

\subsection{Representational Validation}\label{sec:rep-valid}
 
The failure taxonomy rests on causal interventions that target specific visual tokens. A potential confound is that interventions might detect surface linguistic differences between clean and corrupted questions rather than genuine changes in visual computation. 
Table~\ref{tab:probing} reports the vision-sensitivity gap for each operation. The operations central to the vision-dependent failure modes---\texttt{verify}, \texttt{relate}, and \texttt{exist}---show substantial positive gaps, supporting that their causal effects reflect visual representations rather than lexical differences. In contrast, \texttt{filter} has the smallest gap ($\delta_{\mathrm{VS}}=0.03$), independently supporting its language-prior classification; \texttt{query} and \texttt{select} are marginal ($\delta_{\mathrm{VS}}=0.06$).

%Table~\ref{tab:probing} reports the vision-sensitivity gap $\delta_\text{VS}$ for each operation.
 
%\textbf{The three operations central to the taxonomy's vision-dependent failure modes all pass the validation threshold($\delta_\text{VS} > 0.05$) with substantial margins}. In each case, probes trained on pairs where the model demonstrably changed its answer achieve near-perfect accuracy, while probes on pairs where the answer was unchanged perform markedly worse. This confirms that the causal signals underlying the reasoning failure, grounding failure, and attribute extraction failure modes reflect changes in visual representation, not lexical cues.
 
%\texttt{filter} fails the threshold ($\delta_\text{VS} = +0.03$), \textbf{providing independent representational evidence for its classification as vision-disconnected}. \texttt{query} and \texttt{select} are marginal ($\delta_\text{VS} = +0.06$). Their high NV baselines (${\sim}0.94$) indicate that these operations produce sufficiently distinctive question structures for probes to detect the counterfactual from text alone, the positive gap nonetheless confirms a residual vision contribution. We exclude \texttt{choose} from the probing analysis because constructing valid counterfactual pairs for binary-choice questions is not feasible (e.g.\ changing ``Is the chain black or yellow?'' to ``Is the chain yellow or black?'') This would alter only the order of options without introducing a genuine semantic contrast.

\begin{table}[t]
\centering
\caption{Counterfactual probing validation. Vision-sensitive
(VS) and vision-insensitive (NV) probe accuracy at best layer.
The VS–NV gap validates whether probes read visual computation or surface text.}
\label{tab:probing}
\small
\begin{tabular}{l ccc l}
\toprule
\textbf{Op.} & \textbf{VS} & \textbf{NV} 
& $\boldsymbol{\delta}_{\textbf{VS}}$ 
& \textbf{Verdict} \\
\midrule
\texttt{verify} & 1.00 & 0.66 & +0.34 & Vision \\
\texttt{relate} & 0.93 & 0.68 & +0.24 & Vision \\
\texttt{exist}  & 0.98 & 0.75 & +0.23 & Vision \\
\texttt{query}  & 1.00 & 0.94 & +0.06 & Marginal \\
\texttt{select} & 1.00 & 0.94 & +0.06 & Marginal \\
\texttt{filter} & 0.82 & 0.78 & +0.03 & Artifact \\
\bottomrule
\end{tabular}
\end{table}

\begin{table*}[t]
\centering
\caption{\textbf{Operation-level causal profile.}
Peak degradation under mean ablation, direct-attention knockout,
and answer-position MLP knockout. Cohen's $d$ compares correct and
incorrect predictions; ***$p<.001$, **$p<.01$, *$p<.05$.}

\label{tab:main-results}
\small
\begin{tabular}{l c cc c cc c cc c l}
\toprule
& & \multicolumn{3}{c}{\textbf{Mean Ablation}} 
& \multicolumn{3}{c}{\textbf{Attn KO}} 
& \multicolumn{3}{c}{\textbf{MLP KO (answer)}} & \\
\cmidrule(lr){3-5} \cmidrule(lr){6-8} \cmidrule(lr){9-11}
\textbf{Operation} & \textbf{Acc.} 
& $\text{GS}_{\checkmark}$ & $\text{GS}_{\times}$ 
& $d$ 
& $\text{KO}_{\checkmark}$ & $\text{KO}_{\times}$ 
& $d$ 
& $\text{KO}_{\checkmark}$ & $\text{KO}_{\times}$ 
& $d$ 
& \textbf{Failure mode} \\
\midrule
\texttt{select} & 60.4\% 
& \textbf{0.975} & 0.489 & \textbf{+0.75}\rlap{***}
& 0.074 & 0.104 & $-$0.19\phantom{***}
& 1.104 & \textbf{1.439} & \textbf{$-$0.80}\rlap{***}
& Grounding \\

\texttt{relate} & 44.8\% 
& 0.894 & \textbf{1.116} & $-$0.17\rlap{**}\phantom{*}
& 0.164 & 0.298 & \textbf{$-$0.32}\rlap{***}
& 1.438 & 1.543 & $-$0.17\rlap{*}\phantom{**} 
& Reasoning \\

\texttt{verify} & 86.4\% 
& 1.134 & 0.995 & +0.12\phantom{***} 
& 0.050 & 0.045 & 0.05 
& 0.956 & \textbf{1.197} & \textbf{$-$0.71}\rlap{***}
& Attr.\ extraction \\

\texttt{query} & 67.4\% 
& 0.918 & 0.759 & +0.15\phantom{***} 
& 0.224 & 0.234 & $-$0.02
& \textbf{1.965} & 1.739 & \textbf{+0.33}\rlap{***}
& Attr.\ extraction \\

\texttt{exist} & 82.5\% 
& \textbf{1.147} & 0.775 & +0.30\phantom{***}
& 0.043 & 0.060 & $-$0.15\rlap{**}\phantom{*} 
& 0.822 & 0.906 & $-$0.27\rlap{*}\phantom{**} 
& Attr.\ extraction \\

\texttt{choose} & 89.4\% 
& 0.310 & 0.281 & +0.08\phantom{***} 
& 0.190 & 0.176 & +0.10\phantom{***}
& 1.744 & 1.830 & $-$0.13\phantom{***} 
& Language prior \\

\texttt{filter} & 90.0\% 
& 0.214 & 0.217 & $-$0.01\phantom{***} 
& 0.1393 & 0.1453 & +0.0659\phantom{***} 
& 1.438 & 1.464 & $-$0.06\phantom{***} 
& Language prior \\
\bottomrule
\end{tabular}
\end{table*}

\subsection{Grounding Failure: \texttt{select}}
 
\texttt{Select} produces the cleanest mechanistic signal in the study. Mean ablation reveals \textbf{a strong positive association between grounding strength and correctness} ($d = +0.75$). Correct answers depend heavily on the target object's visual representation while incorrect answers show minimal dependence. The model fails because it never develops causal reliance on the target object and the bottleneck is perceptual.
 
The pathway analysis localises this failure to the feedforward network. Attention knockout produces no significant correctness discrimination showing that blocking the answer token's attention to the object barely affects the output regardless of correctness. In contrast, \textbf{MLP knockout at the answer position yields the largest negative effect among all operations} ($d = -0.80$), with incorrect answers substantially more sensitive to MLP disruption. The dissociation is visible in Figure~\ref{fig:pathway-dissociation} (top left), where the full-state ablation gap is not reproduced by direct answer-to-object attention knockout, whereas a strong correctness-dependent effect appears under answer-position MLP knockout.
The \texttt{select} effect remains significant after both Bonferroni and BH-FDR correction ($p<10^{-4}$).
 
\subsection{Reasoning Failure: \texttt{relate}}\label{sec:reasoning}

\texttt{Relate} exhibits the opposite grounding profile. Mean 
ablation shows \textbf{a reversed grounding--correctness 
relationship}. Incorrect answers are more causally dependent on the target object than correct 
ones. The model locates the object but persists in retrieving 
its visual representation at processing stages where abstract 
relational inference should dominate.

\begin{figure}[t]
\centering
\includegraphics[width=\columnwidth]{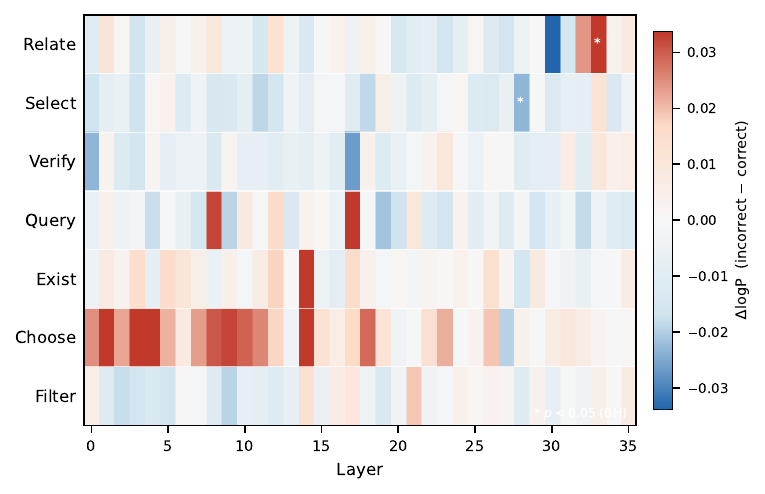}
\caption{Layer-wise attention knockout difference ($\Delta$KO = incorrect $-$ correct) across operations. Red indicates incorrect predictions depend more on bbox attention at that layer; blue indicates correct predictions do. \texttt{relate} shows a concentrated positive signal at L33, confirming attention-mediated over-grounding at late layers.}
\Description{Heatmap of attention-knockout differences across
36 layers and seven operations, with the strongest late-layer
signal for relate near layer 33.}
\label{fig:attn-heatmap}
\end{figure}

For multi-step \texttt{relate} questions in GQA, the strongest correctness effect occurs under direct answer-to-object attention knockout. It \textbf{amplifies the reversed 
signal}, nearly doubling the mean 
ablation effect, with degradation concentrated at layers 28--35 
(Figure~\ref{fig:pathway-dissociation}, top right; 
Figure~\ref{fig:attn-heatmap}). 
The layer-33 peak replicates across two independent runs, and per-head knockout shows that the effect is distributed across multiple heads rather than driven by a single isolated head.
The answer-position MLP 
contributes a secondary effect of similar magnitude to the 
ablation but substantially weaker 
than the attention pathway, indicating that \textbf{the smaller MLP effect indicates secondary feedforward involvement, although the present diagnosis does not establish whether this contribution compensates for or propagates the attention-related signal}.

With the lowest accuracy of all operations (45\%), 
\texttt{relate} is the only failure mode where the model's 
\emph{access} to the correct visual information actively 
harms performance.
The \texttt{relate} effect likewise remains significant after Bonferroni ($p=0.016$) and BH-FDR ($p=0.008$) correction.

\subsection{Attribute Extraction Failure: \texttt{verify}, 
\texttt{query}, \texttt{exist}}

Three operations share a common profile under mean 
ablation: high absolute grounding strength with no 
significant correctness discrimination. \textbf{The model 
consistently grounds the relevant object regardless of 
whether it answers correctly}. Attention knockout confirms 
this pattern as they don't show a 
meaningful correctness split through the attention pathway(Figure~\ref{fig:pathway-dissociation}, bottom left). As a positional control, MLP knockout at intermediate text positions produces no significant correctness-dependent effect for \texttt{verify}, \texttt{query}, or \texttt{exist} ($d=-0.18$, $+0.12$, and $-0.09$, respectively), compared with significant answer-position effects ($d=-0.71$, $+0.33$, and $-0.27$). For \texttt{query}, degradation from the single answer-position knockout (1.77) also exceeds the cumulative degradation across all 13 intermediate positions (1.08), supporting the answer position as the dominant locus.

\textbf{MLP knockout at the answer position reveals the dominant failure 
locus}. For \texttt{verify}, incorrect answers are 
substantially more sensitive to MLP disruption, the second-largest effect after \texttt{select}. 
The feedforward network at the answer position fails to 
extract the correct verification judgment from a 
well-grounded representation even when the model grounds both 
objects in a relation correctly but cannot compute whether 
the relation holds. Object-position MLP knockout 
corroborates this and shows significant 
sensitivity at the bbox positions as well ($d = -0.49$), 
indicating that the MLP's processing of the object 
representation itself also contributes to the failure.

\texttt{Query} displays a distinct variant: the MLP 
effect reverses direction ($d = +0.33$), with correct 
answers more dependent on answer-position MLP 
computation. When the MLP successfully processes the 
grounded representation, the attribute is retrieved 
correctly; although when the MLP does not engage productively, the 
answer defaults to an incorrect response. The failure is 
not that the MLP breaks but that it was never supplied 
with a representation from which the correct attribute 
could be extracted.

\texttt{Exist} shows significant effects in both the object-position MLP ($d = -0.66$) and the attention pathway, with a moderate answer-position MLP effect. Unlike \texttt{verify} and \texttt{query}, \textbf{exist failures are distributed across pathways with no single dominant mechanism}, consistent with existence checking being a simpler operation that can fail at multiple stages. Accordingly, the attribute-extraction label for \texttt{exist} summarizes its dominant operation-level profile rather than an exclusive answer-position mechanism.

%%% ─── LANGUAGE PRIOR ──────────────────────────────────────────
\subsection{Language Prior Dominance: \texttt{choose}, \texttt{filter}}
The \texttt{choose} and \texttt{filter} operations show \textbf{uniformly null results across all three interventions}. Causal mean ablation produces the lowest grounding strength of any operations with no correctness discrimination. Neither attention knockout nor MLP knockout reveals a significant pathway-specific effect. These operations achieve high accuracy (89--90\%) through linguistic patterns rather than visual computation (Figure~\ref{fig:pathway-dissociation}, bottom right).
 
\texttt{Choose} exploits world knowledge encoded in the language model's parameters, it can preserve a substantial fraction of its performance using linguistic and world-knowledge priors when visual evidence is degraded, consistent with the well-documented tendency of VLMs to rely on language priors when the question structure permits~\cite{agrawal2018don}. \texttt{Filter} applies attribute-based subsetting that the model resolves from question semantics alone. Independent evidence detected by linear probing presented in Section \ref{sec:rep-valid} corroborates this classification: \texttt{filter} shows the lowest vision-sensitive pair rate (27\%) and a negligible probe validation gap.
A pixel-level corruption control provides independent evidence for this interpretation. When images are replaced by solid gray or Gaussian noise, \texttt{filter} and \texttt{choose} retain 66.4\% and 51.1\% of their original accuracy, respectively, compared with 43.3\% for \texttt{verify} and 32.0\% for \texttt{relate}. The language-prior operations therefore preserve substantially more performance when visual evidence is removed, although they are not entirely image-independent.
The null correctness association for \texttt{filter} is preserved after Bonferroni ($p=0.637$) and BH-FDR ($p=0.212$) correction.

\subsection{Spatial Relation Validation on VSR}

The GQA and VSR results distinguish two forms of relational failure. GQA \texttt{relate} occurs within a multi-step functional program and shows its strongest correctness-discriminating effect under late-layer direct-attention knockout. VSR requires a single spatial judgment between two explicitly referenced objects and instead shows stronger object-position MLP effects. Together, these results indicate that single-step spatial failures are associated primarily with encoding object-specific spatial representations, whereas multi-step relational composition additionally depends on late-stage access to grounded object information. Table~\ref{tab:vsr_summary} reports the causal profile across all three analyses for each VSR meta-category.

Object-position MLP knockout produces significant correctness-dependent effects in four of seven categories, with the strongest effects for Topological and Proximity relations. Direct-attention effects are limited to Adjacency and Directional relations, where correct predictions depend more strongly on access to the grounded anchor object. Answer-position MLP effects are null across all categories. These results suggest an encoding--extraction distinction: single-step spatial failures are dominated by object-position processing, whereas GQA attribute-extraction failures arise predominantly during answer formation.

\begin{table}[t]
\centering
\caption{\textbf{VSR results by spatial-relation category.}
Cohen's $d$ between correct and incorrect predictions;
***$p<.001$, **$p<.01$, *$p<.05$; ns denotes non-significance.}
\label{tab:vsr_summary}
\footnotesize
\setlength{\tabcolsep}{5pt}
\renewcommand{\arraystretch}{1.0}
\begin{tabular}{l r r r}
\toprule
\textbf{Category} & \textbf{Abl.} & \textbf{Attn} &
\textbf{MLP(obj.)} \\
\midrule
Projective  & $-$.35*** & $-$.13\textsuperscript{ns} & $-$.60*** \\
Topological & $-$.08*   & +.01\textsuperscript{ns}   & $-$.81*** \\
Directional & +.14\textsuperscript{ns} & +.45* & $-$.61** \\
Adjacency   & $-$.23* & +.10* & $-$.30\textsuperscript{ns} \\
Proximity   & $-$.07\textsuperscript{ns} & +.04\textsuperscript{ns} & $-$.72* \\
Orientation & $-$.33\textsuperscript{ns} & $-$.01\textsuperscript{ns} & $-$.03\textsuperscript{ns} \\
Unallocated & +.83* & +.49* & $-$.31\textsuperscript{ns} \\
\bottomrule
\end{tabular}
\end{table}

\subsection{Cross-Architecture Analysis}

The taxonomy is derived primarily from Qwen2.5-VL, whose native-resolution vision encoder preserves spatial detail in the token representation. To examine which profiles transfer across architectures, we apply the same interventions to LLaVA-1.5-7B, which uses a frozen CLIP-ViT encoder and a lightweight two-layer MLP adapter.

Attribute-extraction effects transfer most clearly. \texttt{query} ($d=+0.48$, $p<0.001$) and \texttt{exist} ($d=-0.60$, $p<0.001$) retain significant answer-position MLP effects. In contrast, \texttt{select} and \texttt{relate} show no significant correctness-dependent effects across the three interventions. This absence is consistent with an encoder-sensitive profile: Qwen's bounding-box tokens are 4.2$\times$ more object-separable than LLaVA's ($0.106$ versus $0.025$), suggesting that ablating a specific object's tokens removes less unique spatial information in LLaVA.

The language-prior operations are not uniformly null on LLaVA. \texttt{choose} shows modest significant effects across the interventions, while \texttt{filter} remains weak under mean ablation and attention knockout but shows an answer-position MLP effect. Their classification is independently supported by the image-corruption control, although their precise intervention profile varies across the two models.

The results support a two-tier tendency that output-side attribute-extraction effects are more stable across the two tested models, whereas object-specific grounding and relational profiles are more sensitive to the spatial specificity preserved by the vision encoder.

\section{Conclusion}

We presented an operation-centric framework that diagnoses compositional VQA failures by reasoning operation and dominant computational locus. Across three causal analyses on Qwen2.5-VL, we identify four failure modes: grounding, reasoning, attribute extraction, and language-prior dominance. Object-selection and attribute-extraction failures are associated primarily with feedforward processing, whereas multi-step relational failures are concentrated in late-layer direct attention. VSR further localizes single-step spatial failures to object-position processing. LLaVA provides the strongest cross-architecture support for attribute extraction, while grounding and relational profiles are more sensitive to encoder-preserved spatial specificity. Language-prior dominance is supported independently by image-corruption controls, although its precise intervention profile varies across the two models.

\paragraph{Limitations.}
Our primary analysis uses a 3B-parameter model on GQA, and high-accuracy operations contain relatively few incorrect samples. The identified loci represent dominant rather than exclusive mechanisms, and the framework is diagnostic rather than corrective.
\begin{acks}
This work was supported by the Singapore Institute of Technology
Ignition Grant (STEM), Grant No.\ IG(S) 3/2024-1079.
\end{acks}
\bibliographystyle{ACM-Reference-Format}
\balance
\bibliography{references}

%\appendix

\end{document}

\title{Supplementary Material: How Do VLMs Fail? Vision-Operation Misalignment in Compositional VQA}

\maketitle
\section{Counterfactual Probing: Dataset Construction 
and Training Details}\label{app:probing}

\subsection{Counterfactual Pair Generation}

We construct contrastive question pairs 
$(q_\text{clean}, q_\text{corrupt})$ from GQA's 
balanced training split using a three-pass pipeline. 
Each pair shares the same image but differs in a 
semantically relevant word, with the corruption 
determined by the operation's program argument 
rather than heuristic text matching.

\paragraph{Pass 1: Rule-based corruption.}
For each question, we extract the terminal operation 
and its program argument from GQA's functional 
annotation. The argument specifies exactly what to 
corrupt:

\begin{itemize}
\item \textbf{\texttt{select}}: The argument contains 
  the target object (e.g.\ \texttt{"fruit (1707662)"}). We replace the object name in the question with an absent object not present in the scene graph, drawn from a pool of 16 categories (e.g.\ \emph{unicorn, dinosaur, lighthouse}).

\item \textbf{\texttt{filter}}: The argument contains 
  the filter value (e.g.\ \texttt{"red"}, 
  \texttt{"left"}). We swap it with an 
  alternative from the same attribute category. 
  Attribute pools: 15 colours, 7 shapes, 12 materials, 
  9 sizes, 6 age terms, plus horizontal and vertical 
  position antonyms (\emph{left}$\leftrightarrow$%
  \emph{right}, \emph{top}$\leftrightarrow$%
  \emph{bottom}). Only pairs where the clean 
  answer is \emph{yes} are retained, with the corrupt 
  answer set to \emph{no}.

\item \textbf{\texttt{relate}}: The argument encodes the 
  relation (e.g.\ \texttt{"container, to the right of, 
  s (123)"}). We flip the spatial relation to its 
  antonym using a 17-entry lookup 
  (\emph{left of}$\leftrightarrow$\emph{right of}, 
  \emph{above}$\leftrightarrow$\emph{below}, 
  \emph{inside}$\leftrightarrow$\emph{outside}, etc.). 
  Only \emph{yes}$\to$\emph{no} flips are retained.

\item \textbf{\texttt{verify}}: The argument is the 
  verified attribute (e.g.\ \texttt{"red"}, 
  \texttt{"round"}). We swap it with an alternative 
  from the same category, or with a relation antonym 
  if the verification targets a spatial property.

\item \textbf{\texttt{exist}}: The argument is 
  uninformative (\texttt{"?"}), so we corrupt the 
  upstream dependency --- the \texttt{select}, 
  \texttt{filter}, or \texttt{relate} step that feeds 
  into the existence check --- using the same rules 
  as above, walking up to two levels in the 
  dependency graph.

\item \textbf{\texttt{choose}}: Valid counterfactual pairs for 
  \texttt{choose} are rare because swapping 
  alternatives changes option order rather than 
  introducing a genuine semantic contrast. We 
  therefore exclude \texttt{choose} from the probing 
  analysis.

\item \textbf{\texttt{query}}: Rule-based corruption 
  is not feasible because it requires identifying an 
  alternative attribute type that the target object 
  possesses in the scene graph. \texttt{query} pairs 
  are deferred to Pass~3.
\end{itemize}

All corruptions are validated through human verification upto 50 samples per operation. The corrupt answer must differ from the clean answer, and operation-specific constraints are enforced (e.g.\ \texttt{exist} and 
\texttt{verify} must produce yes/no pairs; 
\texttt{query} must not produce yes/no answers).

\begin{figure*}[t]
\centering
\includegraphics[width=0.7\textwidth]{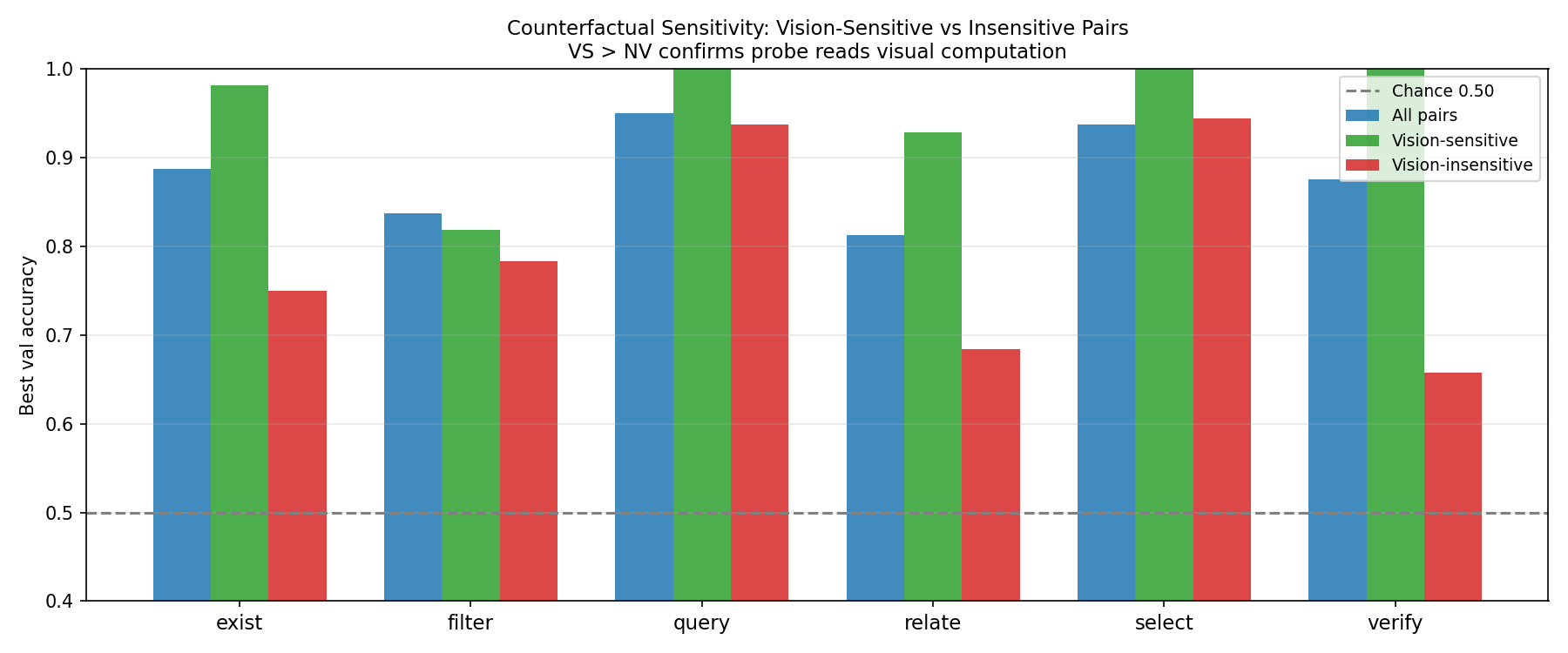}
\caption{VS-NV sensitivity per operation}
\label{fig:probe}
\end{figure*}

\paragraph{Pass 2: Image-centric selection.}
To ensure visual diversity, we cap samples at 3 per image per operation. Images are ranked by the number of distinct operations they cover, and samples are drawn greedily until the target count (200 per operation) is reached.

\paragraph{Pass 3: LLM fallback.}
For operations below the target count after Passes~1--2 (primarily \texttt{query}), we use LLaMA-3.1-70B-Instruct via the NVIDIA NIM API ($T{=}0.2$, top-$p{=}0.7$) to generate corruptions. The LLM receives the scene graph summary, the full 
semantic program, the original question and answer, 
and explicit instructions for what to corrupt. 
Generated pairs are subject to the same validity 
checks as rule-based pairs. For \texttt{query}, the 
LLM is instructed to change the queried attribute 
type (e.g.\ colour$\to$material) and provide the 
correct answer for the new attribute from the scene 
graph.

\paragraph{Dataset statistics.}
Table~\ref{tab:probe-dataset} reports the final 
dataset composition. A total of 1{,}148 pairs across 
six operations are used, with LLM-generated pairs 
comprising the majority of \texttt{query} samples. 
The vision-sensitive rate varies substantially across 
operations, from 27\% for \texttt{filter} to 74\% 
for \texttt{relate}, reflecting the degree to which 
each operation's answer depends on visual computation.
Figure~\ref{fig:probe} shows the Counterfactual Sensitivity between VS and NV pairs.

\begin{table}[h]
\centering
\caption{\textbf{Counterfactual probing dataset.} 
Number of pairs per operation, generation method, and 
vision-sensitivity rate.}
\label{tab:probe-dataset}
\small
\begin{tabular}{l r r r r}
\toprule
\textbf{Op.} & \textbf{n pairs} 
& \textbf{Rule} & \textbf{LLM} 
& \textbf{VS rate} \\
\midrule
\texttt{select} & 200 & 200 & 0   & 44\% \\
\texttt{filter} & 200 & 200 & 0   & 27\% \\
\texttt{relate} & 200 & 200 & 0   & 74\% \\
\texttt{verify} & 200 & 200 & 0   & 58\% \\
\texttt{exist}  & 200 & 200 & 0   & 63\% \\
\texttt{query}  & 148 & 0   & 148 & 52\% \\
\bottomrule
\end{tabular}
\end{table}

\subsection{Probe Training}

We train a single-layer linear probe at each of the 36 transformer layers. 
Input: the hidden state at the last token position 
(answer position) for each question. Output: binary 
classification (clean vs.\ corrupted).

\paragraph{Training details.}
\begin{itemize}
    \item Optimiser: AdamW, $\text{lr}{=}10^{-3}$, weight decay $10^{-4}$
    \item Batch size: 32
    \item Maximum epochs: 200, with early stopping (patience = 5 evaluation rounds, evaluated every 10 epochs)
    \item Train/val split: 80/20, stratified, split at the \emph{pair level} — both members of a pair (clean and corrupted) are assigned to the same split to prevent information leakage
    \item Random seed: 42
\end{itemize}

\subsection{Text-Only Control}

\begin{figure}[h]
\centering
\includegraphics[width=\columnwidth]{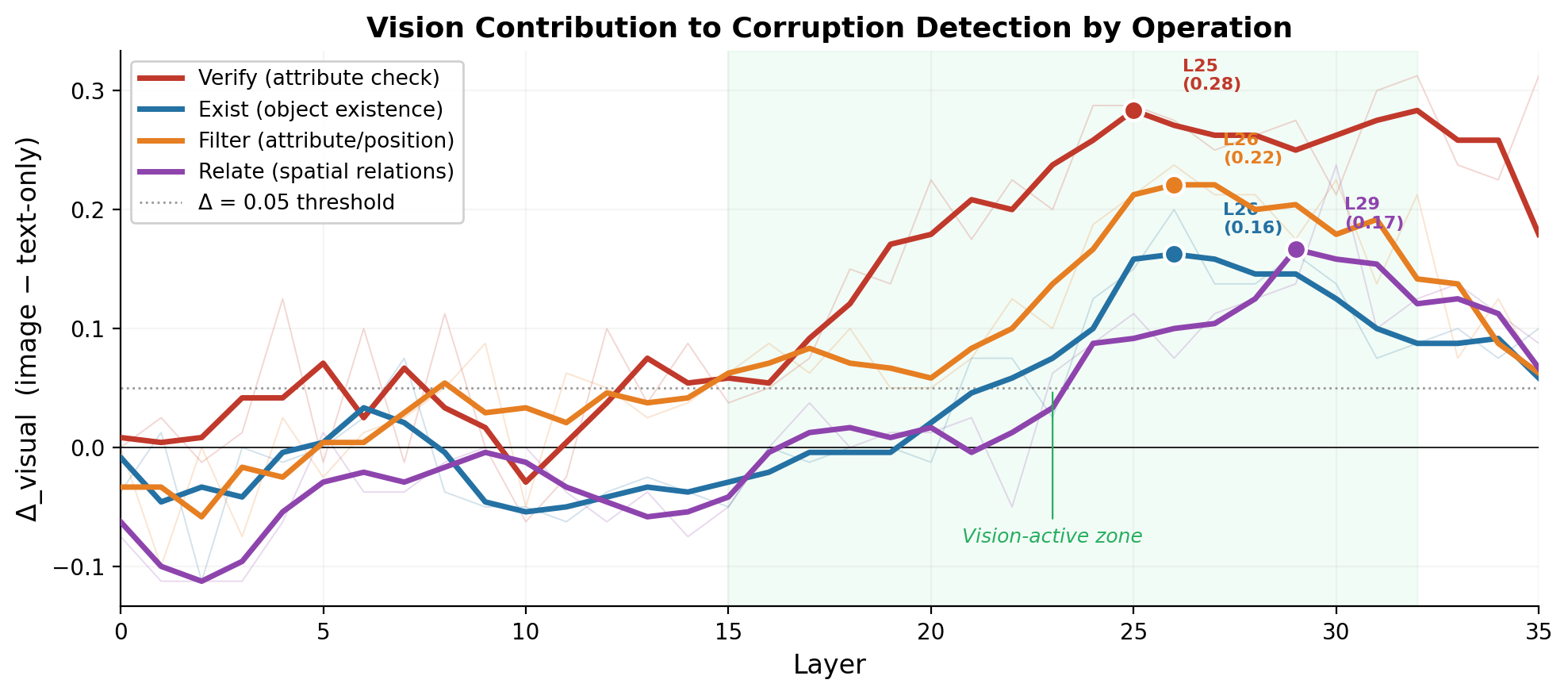}
\caption{\textbf{Vision contribution to corruption 
detection.} $\Delta_\text{visual}$ 
The green zone marks where vision signal exceeds 
the 0.05 threshold.}
\label{fig:delta-visual}
\end{figure}

As a methodological validation, we repeat the 
contrastive probing experiment without images: 
hidden states are extracted from the same questions 
processed through the model's text-only pathway. 
The vision contribution is quantified as 
$\Delta_\text{visual}(L) = \text{acc}_\text{img}(L) 
- \text{acc}_\text{text}(L)$ per layer. 
Figure~\ref{fig:delta-visual} shows the layer-wise 
$\Delta_\text{visual}$ across operations. Vision 
contribution emerges at layers 15--20 and peaks at 
layers 25--29, with \texttt{verify} showing the 
strongest vision signal ($\Delta_\text{max} = 0.28$ 
at L25) and \texttt{relate} showing the latest 
emergence (L29). This temporal ordering is consistent 
with the causal intervention findings: operations 
requiring deeper reasoning show later vision 
integration.

\section{Per-Head Attention Knockout}\label{app:head_knockout}

The layer-level knockout in the main paper treats all 
16 heads at each layer as a single unit. To identify 
which specific heads drive the visual grounding 
effects, we run per-head knockout. For each of the 
$36 \times 16 = 576$ (layer, head) pairs independently in Qwen2.5-VL-3B Instruct, 
we zero out the answer token's attention to the 
bounding-box tokens through that single head and 
measure the resulting degradation. This requires 576 
forward passes per sample.

\subsection{Top Visual Grounding Heads}

Table~\ref{tab:topheads} ranks the five heads with 
the highest mean knockout degradation on correct 
samples for each operation. Two heads emerge as 
\textbf{universally critical}: \textbf{L24H1} appears 
in the top-5 for five of seven operations (filter, 
query, choose, relate, exist) with degradation 
reaching 1.216 nats for query --- the single largest 
per-head effect in the study. \textbf{L23H10} appears 
in the top-5 for all seven operations, making it the 
most consistently important head across all operation 
types. Together, these two heads form a shared visual 
grounding circuit that operates independently of task 
semantics.

Beyond these universal heads, operation-specific 
patterns emerge. \texttt{verify} and \texttt{select} 
are dominated by \textbf{L26H8} and \textbf{L26H3}, 
a pair of mid-to-late layer heads that do not appear 
prominently for other operations. \texttt{filter} 
uniquely relies on \textbf{L27H5} and 
\textbf{L32H6}. \texttt{exist} depends most heavily 
on \textbf{L23H10} and \textbf{L8H1}, the latter 
being an early-layer head that also appears for 
verify and select but not for query or choose.

\begin{table}[h]
\centering
\caption{\textbf{Top-5 heads by mean knockout 
degradation on correct samples} (nats). Each entry 
specifies layer (L) and head index (H). Bold = 
heads appearing in top-5 for $\geq$5 operations.}
\label{tab:topheads}
\small
\setlength{\tabcolsep}{3pt}
\begin{tabular}{l lllll}
\toprule
\textbf{Op.} & \textbf{\#1} & \textbf{\#2} 
& \textbf{\#3} & \textbf{\#4} & \textbf{\#5} \\
\midrule
\texttt{filter} 
& \textbf{L24H1}  \scriptsize{1.04}
& L35H9  \scriptsize{0.81}
& L27H5  \scriptsize{0.61}
& L27H8  \scriptsize{0.59}
& \textbf{L23H10} \scriptsize{0.58} \\
\texttt{relate} 
& \textbf{L24H1}  \scriptsize{0.72}
& \textbf{L23H10} \scriptsize{0.50}
& L26H3  \scriptsize{0.36}
& L26H8  \scriptsize{0.32}
& L35H9  \scriptsize{0.32} \\
\texttt{verify} 
& L26H8  \scriptsize{0.54}
& L26H3  \scriptsize{0.42}
& L8H1   \scriptsize{0.35}
& L35H12 \scriptsize{0.35}
& \textbf{L24H1}  \scriptsize{0.28} \\
\texttt{select} 
& L26H8  \scriptsize{0.67}
& L26H3  \scriptsize{0.54}
& \textbf{L23H10} \scriptsize{0.49}
& L35H12 \scriptsize{0.42}
& L8H1   \scriptsize{0.35} \\
\texttt{query} 
& \textbf{L24H1}  \scriptsize{1.22}
& \textbf{L23H10} \scriptsize{0.75}
& L27H8  \scriptsize{0.65}
& L0H10  \scriptsize{0.62}
& L35H9  \scriptsize{0.54} \\
\texttt{exist} 
& \textbf{L23H10} \scriptsize{0.45}
& L8H1   \scriptsize{0.36}
& L35H12 \scriptsize{0.29}
& L26H13 \scriptsize{0.27}
& \textbf{L24H1}  \scriptsize{0.22} \\
\texttt{choose} 
& \textbf{L24H1}  \scriptsize{1.00}
& \textbf{L23H10} \scriptsize{0.77}
& L27H8  \scriptsize{0.77}
& L35H9  \scriptsize{0.63}
& L32H6  \scriptsize{0.51} \\
\bottomrule
\end{tabular}
\end{table}

\subsection{Per-Head Correctness Dissociation}

Table~\ref{tab:perhead} reports Cohen's $d$ between 
correct and incorrect samples for each of the 16 head 
indices, computed on peak knockout degradation per 
sample (maximum across all 36 layers for that head). 
Negative $d$ indicates incorrect predictions depend 
more on that head's bbox attention.

\texttt{query} is the only operation where 
\textbf{15 of 16 heads show positive $d$}: correct 
answers benefit from bbox attention at virtually 
every head, a unique signature indicating that visual 
grounding is uniformly productive for attribute 
retrieval.

\texttt{exist} shows the most extreme polarisation: 
\textbf{13 of 16 heads are strongly negative}, with 
three heads exceeding $|d| > 1.0$ (H3 $= -2.42$, 
H8 $= -1.34$, H13 $= -1.68$). Incorrect exist 
predictions are dramatically over-reliant on per-head 
bbox attention across the board, not at isolated heads.

\texttt{select} presents a mixed picture: 
\textbf{14 of 16 heads are negative} but one head 
(H3, $d = +0.40$) specifically benefits correct 
predictions, suggesting a partially functional 
grounding circuit rather than wholesale failure.

\texttt{verify} shows \textbf{14 of 16 heads 
negative} (H8 $= -1.06$ most extreme), paralleling 
exist in direction but with smaller magnitudes.

\texttt{relate}, \texttt{choose}, and \texttt{filter} 
show near-zero mixed-polarity patterns with no 
systematic head-level bias, consistent with their 
weaker aggregate attention knockout effects in the 
main results.

\begin{table*}[t]
\centering
\caption{\textbf{Per-head Cohen's $d$ between correct 
and incorrect samples.} Each cell shows $d$ computed 
on peak knockout degradation (max across all 36 
layers) per sample for that head index. Negative 
$d$ = incorrect answers depend more on that head's 
bbox attention. \textbf{Bold} = $|d| > 0.5$.}
\label{tab:perhead}
\small
\setlength{\tabcolsep}{3.5pt}
\begin{tabular}{l rrrrrrrrrrrrrrrr}
\toprule
& H0 & H1 & H2 & H3 & H4 & H5 & H6 & H7 
& H8 & H9 & H10 & H11 & H12 & H13 & H14 & H15 \\
\midrule
\texttt{select} 
& $-$.34 & $-$.46 & \textbf{$-$.75} & +.40 
& \textbf{$-$.85} & \textbf{$-$.51} 
& \textbf{$-$.50} & \textbf{$-$.61}
& \textbf{$-$.52} & \textbf{$-$.65} 
& \textbf{$-$.51} & \textbf{$-$.57}
& $-$.17 & $-$.07 & \textbf{$-$.50} & $-$.04 \\
\texttt{relate} 
& $-$.08 & $-$.09 & +.07 & +.04 
& $-$.30 & $-$.02 & $-$.06 & $-$.13
& +.14 & +.07 & $-$.19 & +.21
& +.07 & +.18 & $-$.04 & $-$.17 \\
\texttt{verify} 
& +.20 & $-$.30 & $-$.33 & \textbf{$-$.75} 
& $-$.13 & $-$.20 & $-$.32 & $-$.28
& \textbf{$-$1.06} & $-$.38 & $-$.35 & +.02
& $-$.22 & \textbf{$-$.71} & $-$.08 
& \textbf{$-$.68} \\
\texttt{query} 
& \textbf{+.53} & +.34 & +.21 & +.36 
& +.19 & +.30 & +.22 & +.22
& +.10 & \textbf{+.63} & +.18 & \textbf{+.65}
& $-$.01 & +.21 & +.33 & +.08 \\
\texttt{exist} 
& \textbf{$-$.80} & $-$.38 & \textbf{$-$.55} 
& \textbf{$-$2.42} 
& $-$.48 & \textbf{$-$.51} & \textbf{$-$.80} 
& $-$.31
& \textbf{$-$1.34} & $-$.44 & +.11 & $-$.46
& +.04 & \textbf{$-$1.68} & $-$.42 
& \textbf{$-$.57} \\
\texttt{choose} 
& $-$.02 & +.02 & +.03 & +.00 
& +.10 & $-$.08 & $-$.03 & +.18
& $-$.14 & +.46 & +.02 & $-$.02
& $-$.18 & $-$.17 & $-$.30 & +.08 \\
\texttt{filter} 
& $-$.11 & +.10 & $-$.16 & $-$.19 
& $-$.13 & +.13 & +.35 & +.04
& $-$.14 & +.32 & $-$.19 & $-$.10
& $-$.39 & $-$.09 & $-$.21 & +.11 \\
\bottomrule
\end{tabular}
\end{table*}

\section{MLP Knockout: Full Results}\label{app:mlp}

The main paper reports answer-position MLP knockout 
(Variant~B). Here we 
present both variants side by side, enabling direct 
comparison of where the feedforward failure localises: 
at the object's representation (Variant~A) or at the 
answer token's computation (Variant~B).

\subsection{Variant A — Object-Position MLP Knockout}

At each layer $L$, the MLP output at bounding-box 
token positions is replaced with the mean MLP output 
across all vision tokens at that layer. This tests 
whether the MLP's transformation of the object's 
representation carries grounding signal, while leaving 
attention routing intact.

\begin{table}[H]
\centering
\caption{\textbf{Object-position MLP knockout (Variant~A).} 
Peak degradation split by correctness. 
Negative $d$ = incorrect predictions are more sensitive to bbox MLP disruption.}
\label{tab:mlp-bbox}
\small
\setlength{\tabcolsep}{5pt}
\begin{tabular}{l rr rr rr}
\toprule
\textbf{Op.} 
& $n_{\checkmark/\times}$ 
& $\text{KO}_{\checkmark/\times}$ 
& $d$ & $p$ 
& $L_{\checkmark/\times}$ \\
\midrule
\texttt{exist}   & 404/96  & 0.134/0.224 & $-$0.66 & $<$0.001 & 11/11 \\
\texttt{verify}  & 433/67  & 0.146/0.227 & $-$0.49 & 0.001    & 20/17 \\
\texttt{choose}  & 447/53  & 0.186/0.158 & $+$0.27 & 0.191    & 18/4  \\
\texttt{relate}  & 221/279 & 0.189/0.227 & $-$0.20 & 0.063    & 30/31 \\
\texttt{filter}  & 451/49  & 0.142/0.130 & $+$0.15 & 0.309    & 16/3  \\
\texttt{query}   & 334/166 & 0.237/0.230 & $+$0.04 & 0.613    & 29/11 \\
\texttt{select}  & 305/195 & 0.202/0.202 & $+$0.01 & 0.477    & 20/9  \\
\bottomrule
\end{tabular}
\end{table}

\textbf{\texttt{exist} shows the strongest 
object-position effect} ($d = -0.66$, 
$p < 0.001$): incorrect existence predictions 
depend 67\% more on the MLP's transformation of 
the bbox tokens, with both correct and incorrect 
peaking at the same layer (L11). Despite zero 
BH-corrected significant layers, the aggregate 
effect is highly significant, indicating the 
difference is consistent but spread across layers 
rather than concentrated at a single depth.

\textbf{\texttt{verify} shows a moderate effect} 
($d = -0.49$, $p = 0.001$) with peak layers 
diverging between correct (L20) and incorrect 
(L17) samples, suggesting the error enters the 
object representation at slightly earlier layers.

\textbf{\texttt{select} is null} ($d = +0.01$): 
the MLP's transformation of the object tokens 
contributes equally to correct and incorrect 
predictions. Combined with the strong 
answer-position effect ($d = -0.80$ in 
Variant~B), this confirms that select's grounding 
failure localises to the answer-position 
computation, not to the quality of the object 
representation itself.

\subsection{Variant B — Answer-Position MLP Knockout}

At each layer $L$, the MLP output at the answer token position is replaced with the mean MLP output across all text token positions. This tests whether the MLP's computation at the readout site differentiates correct from incorrect predictions.

\subsection{Cross-Variant Comparison}

Table~\ref{tab:mlp-cross} summarises the directional 
pattern across both variants for each operation. The 
comparison reveals where the feedforward failure 
localises for each failure mode.

\begin{table}[H]
\centering
\caption{\textbf{Cross-variant MLP comparison.} 
Cohen's $d$ for object-position (A) and 
answer-position (B) knockout. Consistent direction 
across variants indicates the MLP failure spans 
both sites; divergent signs indicate a localised 
failure.}
\label{tab:mlp-cross}
\small
\begin{tabular}{l r r l}
\toprule
\textbf{Op.} & \textbf{A (bbox)} 
& \textbf{B (answer)} & \textbf{Pattern} \\
\midrule
\texttt{verify} & $-$0.49** & $-$0.71*** 
& Both levels \\
\texttt{exist}  & \textbf{$-$0.66}*** & $-$0.27* 
& Bbox-dominant \\
\texttt{select} & +0.01 & \textbf{$-$0.80}*** 
& Answer-only \\
\texttt{relate} & $-$0.20 & $-$0.17* 
& Weak, consistent \\
\texttt{query}  & +0.04 & \textbf{+0.33}** 
& Answer-only (+) \\
\texttt{choose} & +0.27 & $-$0.13 
& Null both \\
\texttt{filter} & +0.15 & $-$0.06 
& Null both \\
\bottomrule
\end{tabular}
\end{table}

Three distinct patterns emerge:

\textbf{Answer-position dominant} (\texttt{select}, 
\texttt{query}): the object-position MLP is null 
while the answer-position MLP shows large effects. 
For \texttt{select}, the object's visual features 
are present in the bbox tokens (Variant~A is null) 
but the answer-position MLP fails to use them 
($d = -0.80$). For \texttt{query}, correct 
predictions depend \emph{more} on the answer MLP 
($d = +0.33$), indicating productive MLP engagement 
when the correct attribute is retrievable.

\textbf{Bbox-position dominant} (\texttt{exist}): 
the strongest effect is at the object tokens 
($d = -0.66$) with a moderate answer-position 
contribution ($d = -0.27$). The error mechanism 
for existence checking is rooted in the MLP's 
transformation of the object representation itself.

\textbf{Both levels} (\texttt{verify}): large 
negative effects at both the bbox ($d = -0.49$) 
and answer ($d = -0.71$) positions. Incorrect 
verify predictions over-rely on feedforward 
processing at every stage --- both the object 
encoding and the answer decoding.

\begin{figure*}[t]
\centering
\includegraphics[width=\textwidth]{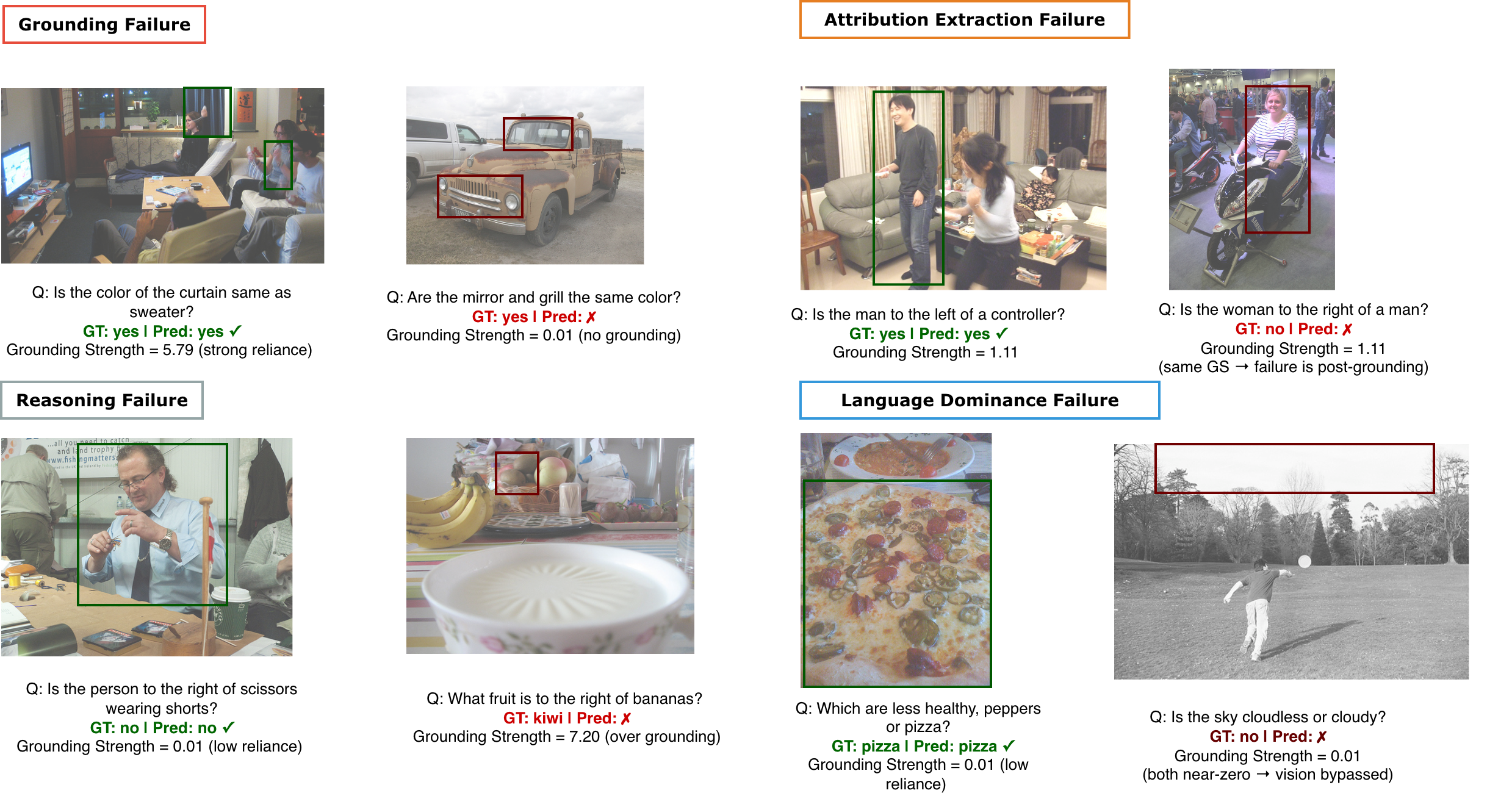}
\caption{\textbf{Qualitative examples by failure 
mode.} Each failure mode is shown with a 
correct (left) and incorrect (right) prediction. 
The GQA image with target object 
bounding box (green = grounded object), the 
question, ground truth, 
and prediction after peak-layer mean ablation. 
The change (or lack thereof) in prediction under 
ablation illustrates the failure mechanism.}
\label{fig:qualitative}
\end{figure*}

\section{Layer-Wise Degradation Curves}
\label{app:layer-curves}

\begin{figure*}[t]
\centering
\includegraphics[width=0.85\textwidth]{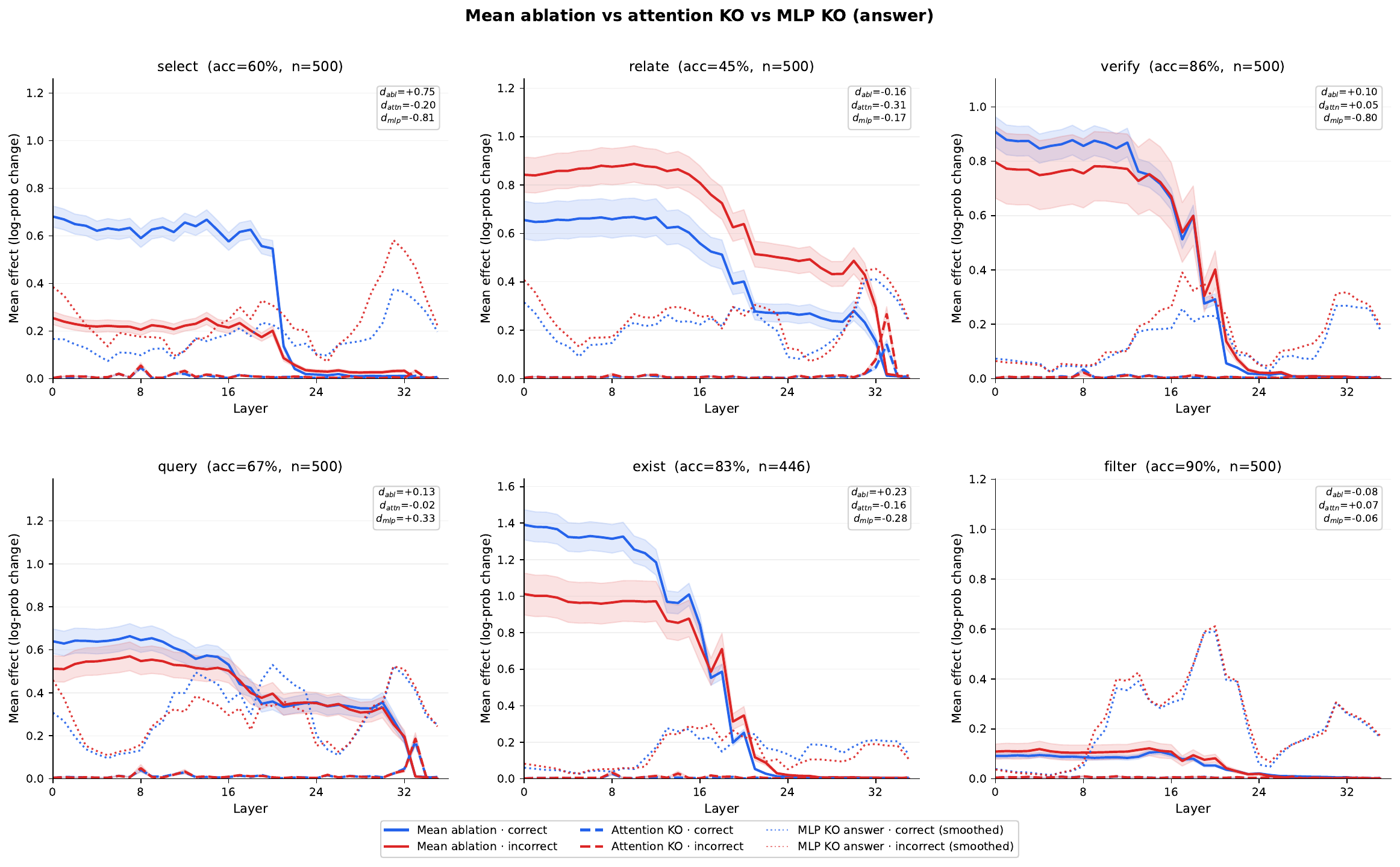}
\caption{\textbf{Layer-wise degradation for all seven 
operations.} Each row is one operation; columns show 
mean ablation, attention knockout, and answer-position 
MLP knockout. Blue = correct predictions, red = 
incorrect. Shading = 95\% CI.}
\label{fig:all-layer-curves}
\end{figure*}

Figure~\ref{fig:all-layer-curves} shows the full 
layer-wise degradation curves for all seven GQA 
operations under each intervention, split by model 
correctness.

\textbf{\texttt{select}}: Mean ablation shows correct 
predictions (blue solid) sustain $\sim$0.6--0.8 nats 
across all layers while incorrect predictions (red 
solid) remain near $\sim$0.2. Attention knockout 
(dashed) is near-zero for both classes across all 
layers, confirming the null attention pathway. MLP 
knockout (dotted) rises sharply at layers 16--24, 
with incorrect predictions (red dotted) reaching 
higher peaks than correct (blue dotted), producing 
the negative $d_\text{mlp}$.

\textbf{\texttt{relate}}: Mean ablation shows 
\emph{reversed} separation where incorrect (red solid) 
runs above correct (blue solid) from layer 8 onward, 
reaching $\sim$0.8--1.0 nats versus $\sim$0.6 for 
correct. Attention knockout shows a sharp spike at 
layers 28--35 where incorrect (red dashed) rises 
above correct (blue dashed), producing the negative 
$d_\text{attn}$ concentrated at late layers. MLP 
knockout curves overlap substantially with a modest 
separation.

\textbf{\texttt{verify}}: Mean ablation curves for 
correct and incorrect track closely together at 
$\sim$0.6--0.8 nats through layers 8--24, with no 
visible separation, it is consistent with the null 
ablation $d$. Attention knockout is near-zero for 
both classes. MLP knockout shows the key signal: 
incorrect (red dotted) rises well above correct 
(blue dotted) from layer 16 onward, peaking at 
$\sim$0.8 versus $\sim$0.5.

\textbf{\texttt{query}}: Mean ablation shows a 
modest correct-above-incorrect separation. Attention 
knockout is negligible. MLP knockout reaches the 
highest absolute values of any operation 
($\sim$1.0--1.2 nats), with correct predictions 
(blue dotted) \emph{above} incorrect (red dotted) 
and the reversed MLP sign unique to query 
($d_\text{mlp} = +0.33$). Correct answers peak 
earlier ($\sim$L21) while incorrect peak later 
($\sim$L33).

\textbf{\texttt{exist}}: Mean ablation shows a 
moderate correct-above-incorrect gap emerging 
around layer 4 and sustained through layer 28, 
reaching $\sim$1.2--1.4 nats for correct versus 
$\sim$0.8 for incorrect. Attention knockout shows 
a small separation with incorrect slightly above 
correct at layers 8--16. MLP knockout curves show 
modest separation with incorrect above correct.

\textbf{\texttt{filter}}: All three interventions 
produce near-identical curves for correct and 
incorrect predictions. Mean ablation reaches 
$\sim$0.6--0.8 nats but with no separation. MLP 
knockout rises at layers 16--24 but with 
overlapping correct and incorrect curves. This 
uniformly null pattern confirms the 
language prior dominance.

\section{Qualitative Examples}\label{app:qualitative}

Figure~\ref{fig:qualitative} presents 
representative examples for each failure mode, 
showing the image with grounded bounding boxes, 
the question, the model's clean prediction, and 
the prediction after intervention at the peak 
degradation layer.

The four rows illustrate the grounding strength (GS) 
signature that defines each failure mode. In 
\textbf{Grounding failure}, the GS gap is extreme: 
the correct prediction relies heavily on the target 
object (GS\,=\,5.79) while the incorrect prediction 
shows near-zero dependence (GS\,=\,0.01). In 
\textbf{Reasoning failure}, the gap reverses: the 
incorrect prediction over-relies on the relational 
endpoint (GS\,=\,7.20) while the correct prediction 
does not (GS\,=\,0.01) showing fixation on the 
object's appearance instead of reasoning about spatial 
context. In \textbf{Attribute Extraction failure}, 
both predictions show matched grounding strength 
(GS\,=\,1.11). The model consistently locates the 
relevant objects but fails downstream when converting 
the grounded representation into the correct 
verification judgment. In \textbf{Language prior dominance}, both predictions are near-zero 
(GS\,=\,0.01) as the model bypasses visual 
computation entirely, answering from language priors 
alone.

\section*{Reproducibility}
Code and samples will be released upon publication.
\begin{table}[H]
\centering
\caption{\textbf{Experimental setup and compute.}}
\label{tab:reproducibility}
\small
\setlength{\tabcolsep}{4pt}
\begin{tabular}{l l}
\toprule
\textbf{Component} & \textbf{Detail} \\
\midrule
\multicolumn{2}{l}{\textit{Hardware}} \\
GPU & NVIDIA H200 140GB $\times$ 1 \\
\midrule
\multicolumn{2}{l}{\textit{Software}} \\
Python & 3.10 \\
PyTorch & 2.2 \\
Transformers & 4.49.0 \\
Precision & fp16 \\
\midrule
\multicolumn{2}{l}{\textit{Compute per experiment}} \\
Mean ablation & 37 fwd passes/sample \\
Attention KO (layer) & 37 fwd passes/sample\\
Attention KO (head) & 576 fwd passes/sample\\
MLP KO (both variants) & 37 fwd passes/sample \\

\bottomrule
\end{tabular}
\end{table}